\documentclass[11pt,a4paper]{article}
\usepackage[hyperref]{acl2020}
\usepackage{times}
\usepackage{latexsym}

\usepackage{microtype}
\usepackage{graphicx}
\usepackage{subcaption}
\usepackage{caption}

\aclfinalcopy 
\def\aclpaperid{8} 

\title{Benchmark and Best Practices for Biomedical Knowledge Graph Embeddings}

\author{David Chang$^{1}$, 
        Ivana Bala\v{z}evi\'c$^{2}$, Carl Allen$^{2}$, Daniel Chawla$^{1}$ \\
        \textbf{Cynthia Brandt}$^{1}$, \textbf{Richard Andrew Taylor}$^{1}$ \\
        $^{1}$Yale Center for Medical Informatics, Yale University \\
        $^{2}$School of Informatics, University of Edinburgh, UK \\
        \texttt{\{david.chang, richard.taylor\}@yale.edu} \\
        \texttt{\{ivana.balazevic, carl.allen\}@ed.ac.uk} 
        }

\date{}

\begin{document}
\maketitle
\begin{abstract}
Much of biomedical and healthcare data is encoded in discrete, symbolic form such as text and medical codes. There is a wealth of expert-curated biomedical domain knowledge stored in knowledge bases and ontologies, but the lack of reliable methods for learning knowledge representation has limited their usefulness in machine learning applications. While text-based representation learning has significantly improved in recent years through advances in natural language processing, attempts to learn biomedical concept embeddings so far have been lacking. A recent family of models called knowledge graph embeddings have shown promising results on general domain knowledge graphs, and we explore their capabilities in the biomedical domain. We train several state-of-the-art knowledge graph embedding models on the SNOMED-CT knowledge graph, provide a benchmark with comparison to existing methods and in-depth discussion on best practices, and make a case for the importance of leveraging the multi-relational nature of knowledge graphs for learning biomedical knowledge representation. The embeddings, code, and materials will be made available to the community\footnote{https://github.com/dchang56/snomed\_kge}.

\end{abstract}

\section{Introduction}
A vast amount of biomedical domain knowledge is stored in knowledge bases and ontologies. For example, SNOMED Clinical Terms (SNOMED-CT)\footnote{https://www.nlm.nih.gov/healthit/snomedct} is the most widely used clinical terminology in the world for documentation and reporting in healthcare, containing hundreds of thousands of medical terms and their relations, organized in a polyhierarchical structure. SNOMED-CT can be thought of as a knowledge graph: a collection of triples consisting of a head entity, a relation, and a tail entity, denoted (h, r, t). SNOMED-CT is one of over a hundred terminologies under the Unified Medical Language System (UMLS) \cite{bodenreider_unified_2004}, which provides a metathesaurus that combines millions of biomedical concepts and relations under a common ontological framework. The unique identifiers assigned to the concepts as well as the Resource Release Format (RRF) standard enable interoperability and reliable access to information. The UMLS and the terminologies it encompasses are a crucial resource for biomedical and healthcare research. 

One of the main obstacles in clinical and biomedical natural language processing (NLP) is the ability to effectively represent and incorporate domain knowledge. A wide range of downstream applications such as entity linking, summarization, patient-level modeling, and knowledge-grounded language models could all benefit from improvements in our ability to represent domain knowledge. While recent advances in NLP have dramatically improved textual representation \cite{alsentzer-etal-2019-publicly}, attempts to learn analogous dense vector representations for biomedical concepts in a terminology or knowledge graph (\textit{concept embeddings}) so far have several drawbacks that limit their usability and wide-spread adoption. Further, there is currently no established best practice or benchmark for training and comparing such embeddings. In this paper, we explore knowledge graph embedding (KGE) models as alternatives to existing methods and make the following contributions:

\begin{itemize}
    \item We train five recent KGE models on SNOMED-CT and demonstrate their advantages over previous methods, making a case for the importance of leveraging the multi-relational nature of knowledge graphs for biomedical knowledge representation.
    \item We establish a suite of benchmark tasks to enable fair comparison across methods and include much-needed discussion on best practices for working with biomedical knowledge graphs.
    \item We also serve the general KGE community by providing benchmarks on a new dataset with real-world relevance.
    \item We make the embeddings, code, and other materials publicly available and outline several avenues of future work to facilitate progress in the field.
\end{itemize}

\section{Related Work and Background}
\subsection{Biomedical concept embeddings}

Early attempts to learn biomedical concept embeddings have applied variants of the skip-gram model \cite{mikolov_2013} on large biomedical or clinical corpora. Med2Vec \cite{choi_multi-layer_2016} learned embeddings for 27k ICD-9 codes by incorporating temporal and co-occurrence information from patient visits. Cui2Vec \cite{beam_clinical_2019} used an extremely large collection of multimodal medical data to train embeddings for nearly 109k concepts under the UMLS. 

These corpus-based methods have several drawbacks. First, the corpora are inaccessible due to data use agreements, rendering them irreproducible. Second, these methods tend to be data-hungry and extremely data inefficient for capturing domain knowledge. In fact, one of the main limitations of language models in general is their reliance on the distributional hypothesis, essentially making use of mostly co-occurrence level information in the training corpus \cite{Peters2019KnowledgeEC}. Third, they do a poor job of achieving sufficient concept coverage: Cui2Vec, despite its enormous training data, was only able to capture 109k concepts out of over 3 million concepts in the UMLS, drastically limiting its downstream usability.

A more recent trend has been to apply network embedding (NE) methods directly on a knowledge graph that represents structured domain knowledge. NE methods such as Node2Vec \cite{node2vec-grover} learn embeddings for nodes in a network (graph) by applying a variant of the skip-gram model on samples generated using random walks, and they have shown impressive results on node classification and link prediction tasks on a wide range of network datasets. In the biomedical domain, CANode2Vec \cite{kotitsas_embedding_2019} applied several NE methods on single-relation subsets of the SNOMED-CT graph, but the lack of comparison to existing methods and the disregard for the heterogeneous structure of the knowledge graph substantially limit its significance. 

Notably, Snomed2Vec \cite{agarwal_snomed2vec_2019} applied NE methods on a clinically relevant multi-relational subset of the SNOMED-CT graph and provided comparisons to previous methods to demonstrate that applying NE methods directly on the graph is more data efficient, yields better embeddings, and gives explicit control over the subset of concepts to train on. However, one major limitation of NE approaches is that they relegate relationships to mere indicators of connectivity, discarding the semantically rich information encoded in multi-relational, heterogeneous knowledge graphs. 

We posit that applying KGE methods on a knowledge graph is more principled and should therefore yield better results. We now provide a brief overview of the KGE literature and describe our experiments in Section 3.

\subsection{Knowledge Graph Embeddings}

Knowledge graphs are collections of facts in the form of ordered triples (\textbf{h}, \textbf{r}, \textbf{t}), where entity \textbf{h} is related to entity \textbf{t} by relation \textbf{r}. Because knowledge graphs are often incomplete, an ability to infer unknown facts is a fundamental task (link prediction). A series of recent KGE models approach link prediction by learning embeddings of entities and relations based on a scoring function that predicts a probability that a given triple is a fact.

RESCAL \cite{rescal2011} represents relations as a bilinear product between subject and object entity vectors. Although a very expressive model, RESCAL is prone to overfitting due to the large number of parameters in the full rank relation matrix, increasing quadratically with the number of relations in the graph. 

DistMult \cite{yang_embedding_2015} is a special case of RESCAL with a diagonal matrix per relation, reducing overfitting. However, by limiting linear transformations on entity embeddings to a stretch, DistMult cannot model asymmetric relations.

ComplEx \cite{trouillon2016complex} extends DistMult to the complex domain, enabling it to model asymmetric relations by introducing complex conjugate operations into the scoring function.

SimplE \cite{kazemi} modifies Canonical Polyadic (CP) decomposition \cite{hitchcock} to allow two embeddings for each entity (head and tail) to be learned dependently. 

A recent model TuckER \cite{balazevic2019tucker} is shown to be a fully expressive, linear model that subsumes several tensor factorization based approaches including all models described above.

TransE \cite{transe} is an example of an alternative \textit{translational} family of KGE models, which regard a relation as a translation (vector offset) from the subject to the object entity vectors. Translational models have an additive component in the scoring function, in contrast to the multiplicative scoring functions of bilinear models. 

RotatE \cite{sun2018rotate} extends the notion of translation to rotation in the complex plane, enabling the modeling of symmetry/antisymmetry, inversion, and composition patterns in knowledge graph relations. 

We restrict our experiments to five models due to their available implementation under a common, scalable platform \cite{zhu2019graphvite}: TransE, ComplEx, DistMult, SimplE, and RotatE.

\section{Experimental Setup}

\subsection{Data}
%
Given the complexity of the UMLS, we detail our preprocessing steps to generate the final dataset. We subset the 2019AB version of the UMLS to \verb SNOMED_CT_US  terminology, taking all active concepts and relations in the MRCONSO.RRF and MRREL.RRF files. We extract semantic type information from MRSTY.RRF and semantic group information from the Semantic Network website\footnote{https://semanticnetwork.nlm.nih.gov} to filter concepts and relations to 8 broad semantic groups of interest: Anatomy (\textbf{ANAT}), Chemicals \&  Drugs (\textbf{CHEM}), Concepts \&  Ideas (\textbf{CONC}), Devices (\textbf{DEVI}), Disorders (\textbf{DISO}), Phenomena (\textbf{PHEN}), Physiology (\textbf{PHYS}), and Procedures (\textbf{PROC}). We also exclude specific semantic types deemed unnecessary. A full list of the semantic types included in the dataset and their broader semantic groups can be found in the Supplements. 

The resulting list of triples comprises our final knowledge graph dataset. Note that the UMLS includes reciprocal relations (\verb ISA  and \verb INVERSE_ISA ), making the graph bidirectional. A random split results in train-to-test leakage, which can inflate the performance of weaker models \cite{dettmers2018conve}. We fix this by ensuring reciprocal relations are in the same split, not across splits. Descriptive statistics of the final dataset are shown in Table~\ref{tab:dataset}. After splitting, we also ensure there are no unseen entities or relations in the validation and test sets by simply moving them to the train set. More details and the code used for data preparation are included in the Supplements.

\begin{table}[h]
\centering
\begin{tabular}{l|r}
\hline
\textbf{Descriptions} & \textbf{Statistics} \\
\hline
Entities & 293,884 \\
Relation types & 170 \\
Facts & 2,073,848 \\ 
 - Train & 1,965,032  \\ 
 - Valid / Test & 48,936 / 49,788  \\\hline
\end{tabular}
\caption{Statistics of the final SNOMED dataset.}\label{tab:dataset}
\end{table}

\subsection{Implementation}
%
Considering the non-trivial size of SNOMED-CT and the importance of scalability and consistent implementation for running experiments, we use GraphVite \cite{zhu2019graphvite} for the KGE models. GraphVite is a graph embedding framework that emphasizes scalability, and its speedup relative to existing implementations is well-documented\footnote{https://github.com/DeepGraphLearning/graphVite}. While the backend is written largely in C++, a Python interface allows customization. We make our customized Python code available. We use the five models available in GraphVite in our experiments: TransE, ComplEx, DistMult, SimplE, and RotatE. While we restrict our current work to these models, future work should also consider other state-of-the-art models such as TuckER \cite{balazevic2019tucker} and MuRP \cite{balazevic2019multi}, especially since MuRP is shown to be particularly effective for graphs with hierarchical structure. Pretrained embeddings for Cui2Vec and Snomed2Vec were used as provided by the authors, with dimensionality 500 and 200, respectively.

All experiments were run on 3 GTX-1080ti GPUs, and final runs took $\sim$6 hours on a single GPU. Hyperparameters were either tuned on the validation set for each model: \verb margin  (4, 6, 8, 10) and \verb learning_rate  (5e-4, 1e-4, 5e-5, 1e-5); set: \verb num_negative  (60), \verb dim  (512), \verb num_epoch  (2000); or took default values from GraphVite. The final hyperparameter configuration can be found in the Appendix.

\subsection{Evaluation and Benchmark}

\subsubsection{KGE Link Prediction}

A standard evaluation task in the KGE literature is link prediction. However, NE methods also use link prediction as a standard evaluation task. While both predict whether two nodes are connected, NE link prediction performs binary classification on a balanced set of positive and negative edges based on the assumption that the graph is complete. In contrast, knowledge graphs are typically assumed incomplete, making link prediction for KGE a ranking-based task in which the model's scoring function is used to rank candidate samples without relying on ground truth negatives. In this paper, link prediction refers to the latter ranking-based KGE method.

Candidate samples are generated for each triple in the test set using all possible entities as the target entity, where the target can be set to \verb head , \verb tail , or \verb both . For example, if the target is \verb tail ,  the model predicts scores for all possible candidates for the tail entity in (h, r, ?). For a test set with 50k triples and 300k possible unique entities, the model calculates scores for fifteen billion candidate triples. The candidates are filtered to exclude triples seen in the train, validation, and test sets, so that known triples do not affect the ranking and cause false negatives. Several ranking-based metrics are computed based on the sorted scores. Note that SNOMED-CT contains a \textit{transitive closure} file, which lists explicit transitive closures for the hierarchical relations \verb ISA  and \verb INVERSE_ISA  (if A \verb ISA  B, and B \verb ISA  C, then the transitive closure includes A \verb ISA  C). This file should be included in the file list used to filter candidates to best enable the model to learn hierarchical structure.

Typical link prediction metrics include Mean Rank (MR), Mean Reciprocal Rank (\textbf{MRR}), and Hits@k (\textbf{H@k}). MR is considered to be sensitive to outliers and unreliable as a metric. \citeauthor{guu-etal-2015-traversing} proposed using Mean Quantile (\textbf{MQ}) as a more robust alternative to MR and MRR. We use MQ$_{100}$ as a more challenging version of MQ that introduces a cut-off at the top 100th ranking, appropriate for the large numbers of possible entities. Link prediction results are reported in Table~\ref{tab:link_prediction}.

\subsubsection{Embedding Evaluation}

For fair comparison with existing methods, we perform some of the benchmark tasks for assessing medical concept embeddings proposed by \citeauthor{beam_clinical_2019}. However, we discuss their methodological flaws in Section \ref{sec:results} and suggest more appropriate evaluation methods.

Since non-KGE methods are not directly comparable on tasks that require both relation and concept embeddings, to compare embeddings across methods we perform entity semantic classification, which requires only concept embeddings. 

We generate a dataset for entity classification by taking the intersection of the concepts covered in all (7) models, comprising 39k concepts with 32 unique semantic types and 4 semantic groups. We split the data into train and test sets with 9:1 ratio, and train a simple linear layer with 0.1 dropout and no further hyperparameter tuning. The single linear layer for classification assesses the linear separability of semantic information in the entity embedding space for each model. Results for semantic type and group classification are reported in Table~\ref{tab:eval_results}.

\section{Visualization}

\begin{figure*}[ht]
\centering
    \includegraphics[width=\linewidth]{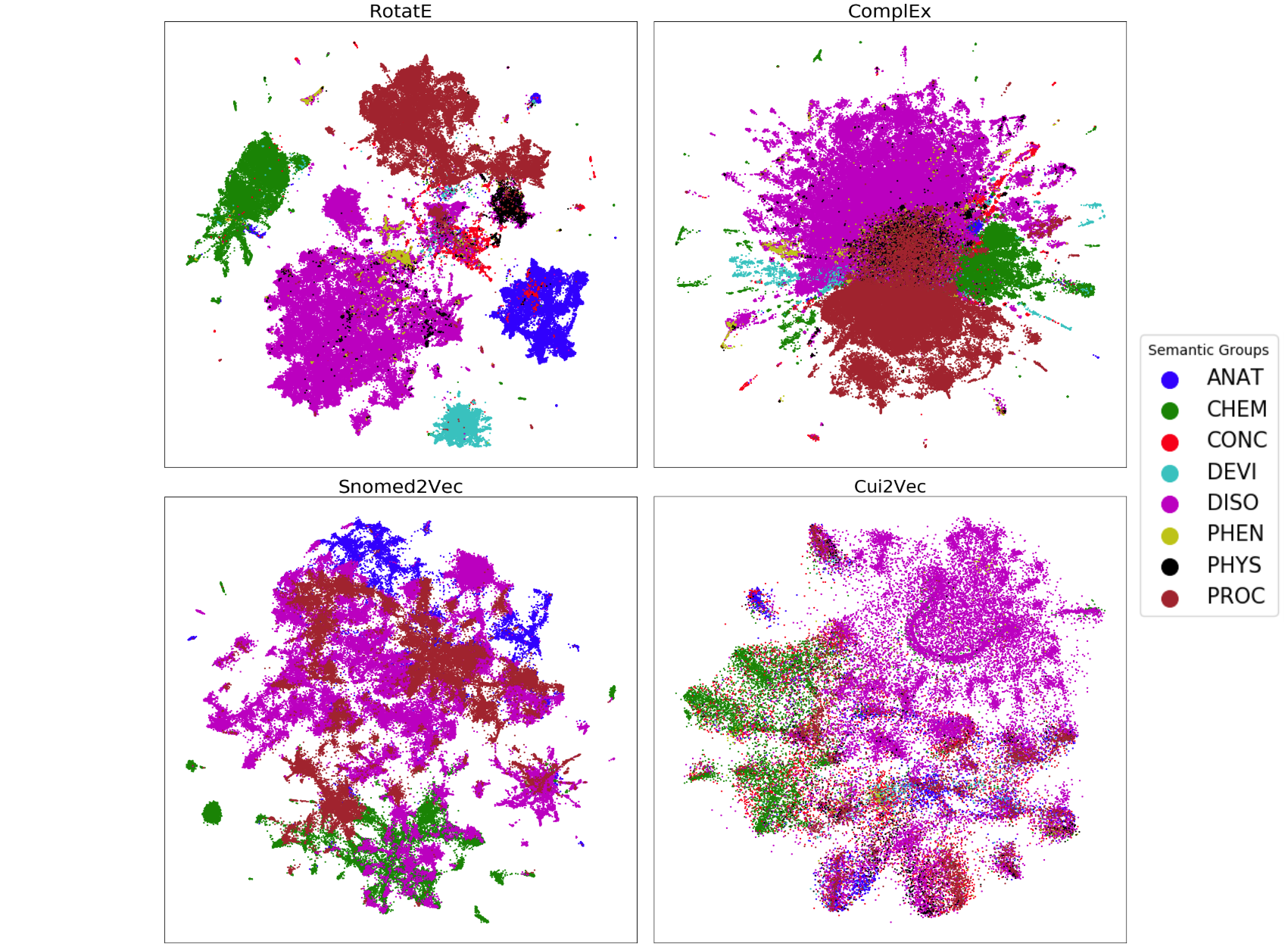}
    \caption{Concept embedding visualization (RotatE, ComplEx, Snomed2Vec, Cui2Vec) by semantic group.}
    \label{fig:broad}
    \vspace{-5pt}
\end{figure*}

\begin{figure*}[ht]
\centering
    \includegraphics[width=0.32\linewidth]{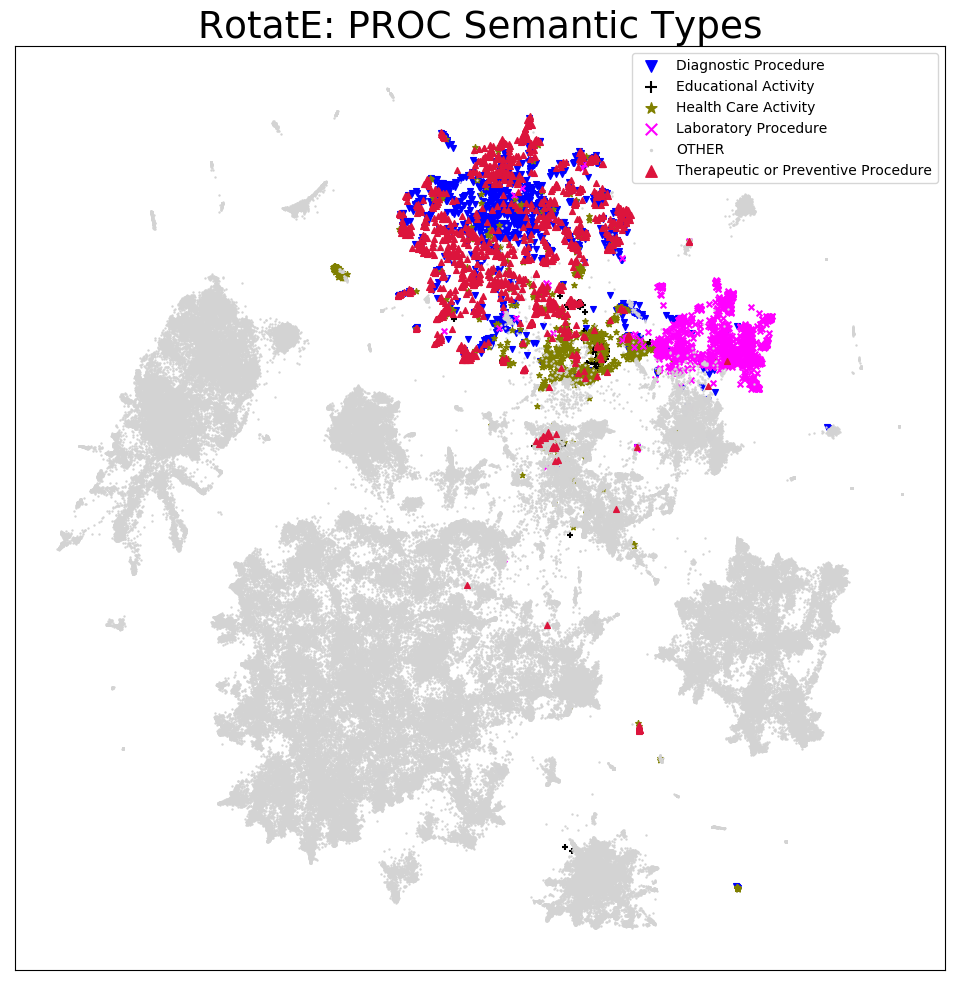}
    \includegraphics[width=0.32\linewidth]{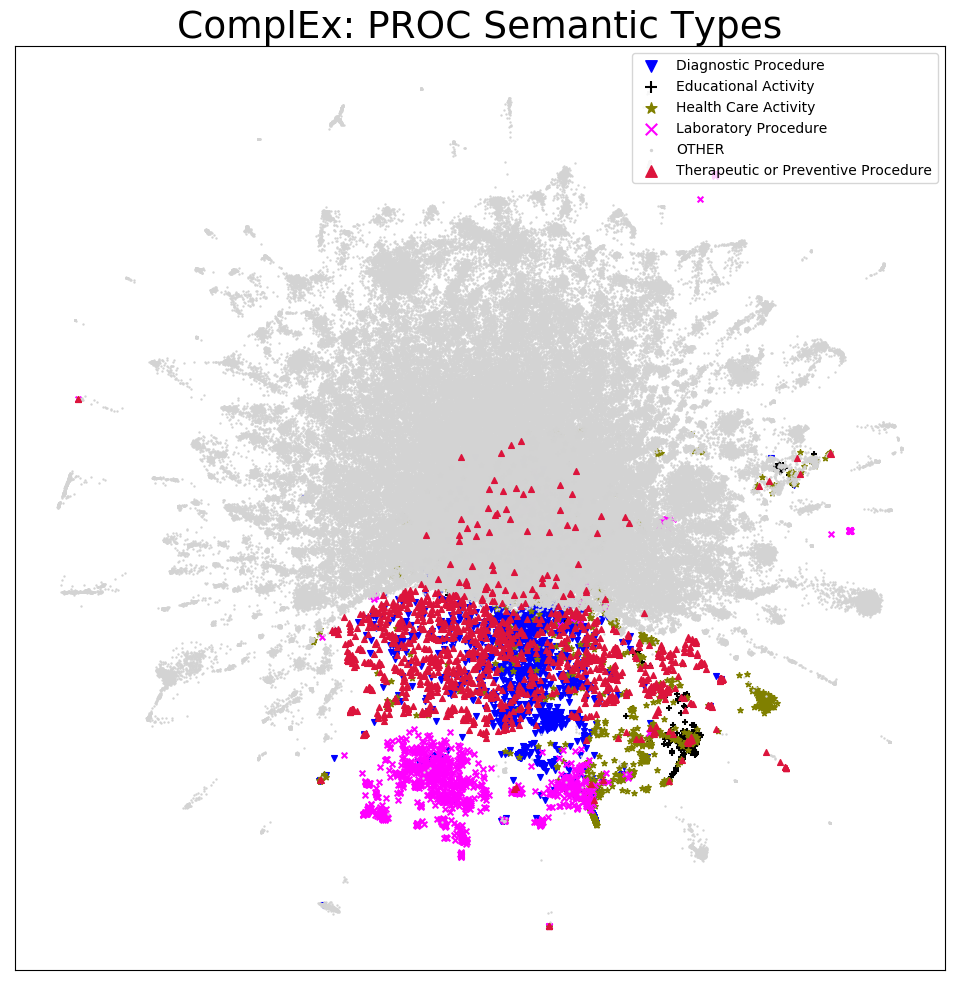}
    \includegraphics[width=0.32\linewidth]{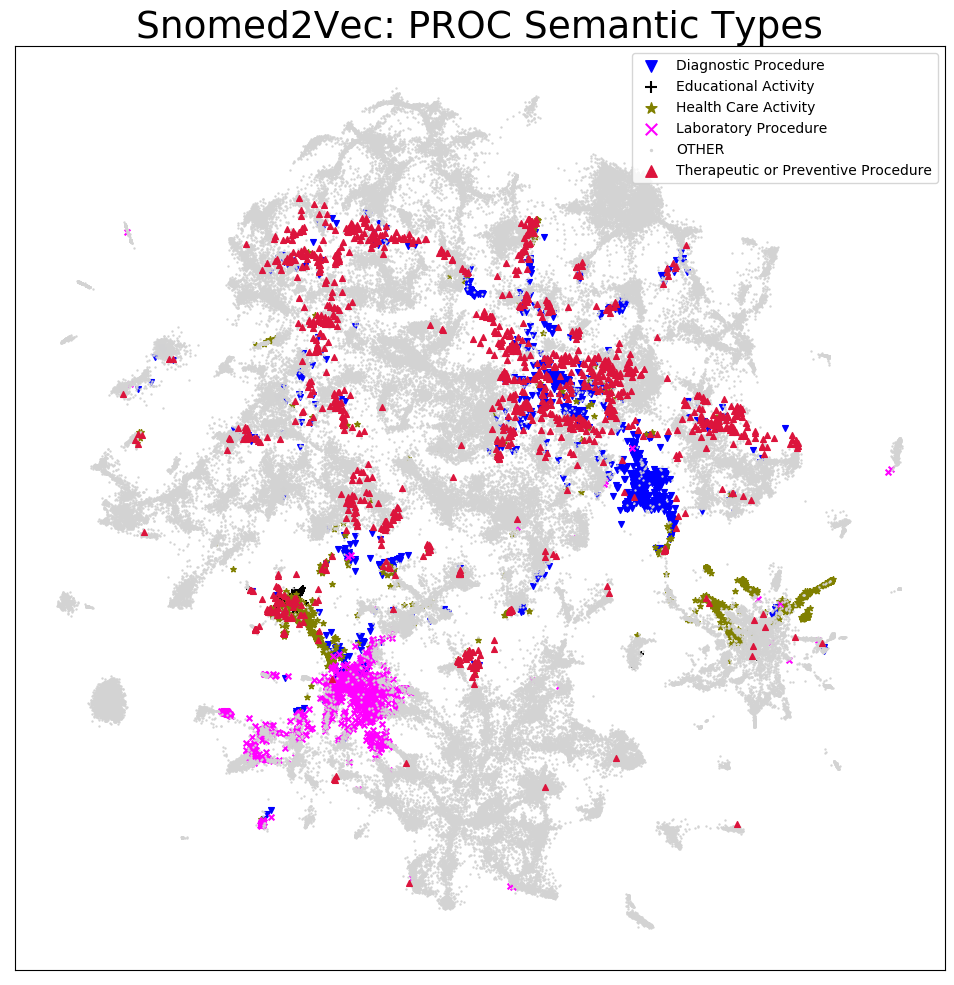}
    \caption{Visualization of selected semantic types under the Procedures semantic group for RotatE, ComplEx, and Snomed2Vec. Semantic types with more than 2,000 entities were subsampled to 1,200 for visibility. Cui2Vec (not shown) was similar to Snomed2Vec but more dispersed.}
    \label{fig:specific}
\end{figure*}

We first discuss the embedding visualizations obtained through LargeVis \cite{tang2016visualizing}, an efficient large-scale dimensionality reduction technique available as an application in GraphVite.

Figure~\ref{fig:broad} shows concept embeddings for RotatE, ComplEx, Snomed2Vec, and Cui2Vec, with colors corresponding to broad semantic groups. Cui2Vec embeddings show structure but not coherent semantic clusters. Snomed2Vec shows tighter groupings of entities, though the clusters are patchy and scattered across the embedding space. ComplEx produces globular clusters centered around the origin, with clearer boundaries between groups. RotatE gives visibly distinct clusters with clear group separation that appear intuitive: entities of the Physiology semantic group (black) overlap heavily with those of Disorders (magenta); also entities under the Concepts semantic group (red) are relatively scattered, perhaps due to their abstract nature, compared to more concrete entities like Devices (cyan), Anatomy (blue), and Chemicals (green), which form tighter clusters. 

Interestingly, the embedding visualizations for the 5 KGE models fall into 2 types: RotatE and TransE produce well-separated clusters while ComplEx, DistMult and SimplE produce globular clusters around the origin. Since the plots for each type appear almost indistinguishable we show one from each (RotatE and ComplEx). We attribute the characteristic difference between the two model types to the nature of their scoring functions: RotatE and TransE have an additive component while ComplEx, DistMult and SimplE are multiplicative. 

Figure~\ref{fig:specific} shows more fine-grained semantic structure by coloring 5 selected semantic types under the Procedures semantic group and greying out the rest. We see that RotatE produces subclusters that are also intuitive. \textit{Laboratory procedures} are well-separated on their own, \textit{health care activity} and \textit{educational activity} overlap significantly, and \textit{diagnostic procedures} and \textit{therapeutic or preventative procedures} overlap significantly. ComplEx also reveals subclusters with globular shape, and Snomed2Vec captures \textit{laboratory procedures} well but leaves other types scattered. These observations are consistent across other semantic groups. We include similar visualizations for the Chemicals \& Drugs semantic group in the Supplements. 

While semantic class information is not the only significant aspect of SNOMED-CT, since the SNOMED-CT graph is largely organized around semantic group and type information, it is promising that embeddings learned (without supervision) preserve it.

\section{Results}
\label{sec:results}

\subsection{Link Prediction}

\begin{table}[ht]
    \centering
    \begin{tabular}{l|c|c|c|c}
    \hline
        \textbf{Model} & \textbf{MRR} & \textbf{MQ$_{100}$} & \textbf{H@1} & \textbf{H@10} \\
    \hline
        TransE & .346 & .739 & .212 & .597 \\
        ComplEx & .461 & .761 & .360 & .652 \\
        DistMult & .420 & .752 & .309 & .626 \\
        SimplE & .432 & .735 & .337 & .615 \\
        RotatE & .317 & .742 & .162 & .599 \\\hline
        TransE\small{$_{FB}$} & .294 & - & - & .465 \\
        TransE\small{$_{WN}$} & .226 & - & - & .501 \\
        RotatE\small{$_{FB}$} & .338 & - & .241 & .533 \\
        RotatE\small{$_{WN}$} & .476 & - & .428 & .571 \\
    \hline
    \end{tabular}
    \caption{Link prediction results: for the 5 KGE models on SNOMED-CT (top); and for TransE and RotatE on two standard KGE datasets \cite{sun2018rotate} (bottom).}
    \label{tab:link_prediction}
\end{table}

Table~\ref{tab:link_prediction} shows results for the link prediction task for the 5 KGE models on SNOMED-CT. Having no previous results to compare to, we include performance of TransE and RotatE on two standard KGE benchmark datasets for reference: FB15k-237 (14,541 entities, 237 relations, and 310,116 triples) and WN18RR (40,943 entities, 11 relations, and 93,003 triples). Given that SNOMED-CT is larger and arguably a more complex knowledge graph than the two datasets, the link prediction results suggest that the KGE models learn a reasonable representation of SNOMED-CT. We include sample model outputs for the top 10 entity scores for link prediction in the Supplements.

\subsection{Embedding Evaluation and Relation Prediction}

\begin{table*}[ht]
    \centering
    \begin{tabular}{l|c|c|c|c|c|c|c|c}
    \hline
          & \multicolumn{2}{|c|}{Entity Classification} & \multicolumn{3}{|c|}{Cosine-Sim Bootstrap} & \multicolumn{3}{|c}{Relation Prediction} \\
    \hline
        \textbf{Model} & \textbf{SG \small{(4)}} & \textbf{STY \small{(32)}} & \textbf{ST} & \textbf{CA} & \textbf{Co} & \textbf{MRR} & \textbf{H@1} & \textbf{H@10} \\
    \hline
        Snomed2Vec   & .944 & .769 & .387 & \textbf{.903} & .894&    - &    - &    - \\
        Cui2Vec      & .891 & .673 & .416 & .584 & .559&    - &    - &    - \\\hline
        TransE       & \textbf{.993} & \textbf{.827} & \textbf{.579} & .765 & \textbf{.978}& .800 & .727 & \textbf{.965} \\
        ComplEx      & .956 & .786 & .249 & .001 & .921& .731 & .606 & .914 \\
        DistMult     & .971 & .794 & .275 & .014 & .971& .734 & .569 & .946 \\
        SimplE       & .953 & .768 & .242 & .011 & .791& \textbf{.854} & \textbf{.803} & .946 \\
        RotatE       & \textbf{.995} & \textbf{.829} & .544 & .242 & .943& \textbf{.849} & \textbf{.799} & .957 \\\hline
    \end{tabular}
    \caption{Results for (i) entity classification of semantic type and group (test accuracy); (ii) selected tasks from \cite{beam_clinical_2019}; and (iii) relation prediction. Best results in bold.}
    \label{tab:eval_results}
\end{table*}

Test set accuracy for entity semantic type (\textbf{STY}) and semantic group (\textbf{SG}) classification are reported in Table~\ref{tab:eval_results}. In accordance with the visualizations of semantic clusters (Figures~\ref{fig:broad} and~\ref{fig:specific}), the KGE and NE methods perform significantly better than the corpus-based method (Cui2Vec). Notably, TransE and RotatE attain near-perfect accuracy for the broader semantic group classification (4 classes). ComplEx, DistMult, and SimplE perform slighty worse, Snomed2Vec slightly below them, and Cui2Vec falls behind by a significant margin. We see a greater discrepancy in relative performance by model type in semantic type classification (32 classes), in which more fine-grained semantic information is required.

Two advantages of the semantic type and group entity classification tasks are: (i) information is provided by the UMLS, making the task non-proprietary and standardized; (ii) it readily shows whether a model preserves the semantic structure of the ontology, an important aspect of the data. The tasks can also easily be modified for custom data and specific domains, e.g. class labels for genes and proteins relevant to a particular biomedical application can be used in classification to assess how well the model captures relevant domain-specific information. 

For comparison to related work, we also examine the benchmark tasks to assess medical concept embeddings based on \textit{statistical power} and \textit{cosine similarity bootstrapping}, proposed by \citeauthor{beam_clinical_2019}. For a given known relationship pair (e.g. x \verb cause_of  y), a null distribution of pairwise cosine similarity scores is computed by bootstrapping 10,000 samples of the same semantic category as x and y respectively. The cosine similarity of the observed sample is compared to the 95th percentile of the bootstrap distribution (statistical significance at the 0.05 level). The authors claim that, when applied to a collection of known relationships (causative, comorbidity, etc), the procedure estimates the fraction of true relationships discovered given a tolerance for some false positive rate. Following this, we report the statistical power of all 7 models for two of the tasks: \textit{semantic type} and \textit{causative relationships}. The former (\textbf{ST}) aims to assess a model's ability to determine if two concepts share the same semantic type. The latter consists of two relation types: \verb cause_of  (\textbf{Co}) and \verb causative_agent_of  (\textbf{CA}). Results are reported in Table~\ref{tab:eval_results}.
The cosine similarity bootstrap results, particularly for the causative relationship tasks, illustrate a major flaw in the protocol. While Snomed2Vec and Cui2Vec attain similar statistical powers for \textbf{CA} and \textbf{Co}, we see large discrepancies between the two tasks for the KGE models, especially for ComplEx, DistMult, and SimplE, which produce globular embedding clusters. Examining the dataset, we observe that the \verb cause_of  relations occur mostly between concepts \textit{within} the same semantic group/cluster (e.g. Disorder), whereas the \verb causative_agent_of  relations occur between concepts in \textit{different} semantic groups/clusters (e.g. Chemicals to Disorders). The large discrepancy in \textbf{CA} task results for the KGE models is because using cosine similarity embeds the assumption that all related entities are close, regardless of the relation type.
The assumption that cosine similarity in the concept embedding space is an appropriate measure of a diverse range of relatedness (a much broader abstraction that subsumes semantic similarity and causality), renders this evaluation protocol unsuitable for assessing a model's ability to capture specific types of relational information in the embeddings. Essentially, all that can be said about the cosine similarity-based procedure is that it assesses how close entities are in that space as measured by cosine distance. It does not reveal the nature of their relationship or what kind of relational information is encoded in the space to begin with.


In contrast, KGE methods explicitly model relations and are better equipped to make inferences about the relational structure of the knowledge graph embeddings. Thus, we propose \textit{relation prediction} as a standard evaluation task for assessing a model's ability to capture information about relations in the knowledge graph. We simply modify the link prediction task described above to accommodate \verb relation  as a target (formulated as (h, ?, t), generating ranking-based metrics for the model's ability to prioritize the correct relation type given a pair of concepts. This provides a more principled and interpretable way to evaluate the models' relation representations directly based on the model prediction. The last 3 columns of Table~\ref{tab:eval_results} report relation prediction metrics for the 5 KGE models. In particular, RotatE and SimplE perform well, attaining around 0.8 Hits@1 and around 0.85 MRR. 

We conduct error analysis to gain further insight by categorizing relation types into 6 groups based on the \textit{cardinality} and \textit{homogeneity} of their source and target semantic groups. If the set of unique head or tail entities for a relation type in the dataset belongs to only one semantic group, then it has a cardinality of 1, and a cardinality of \verb many  otherwise. If the mapping of the source semantic groups to the target semantic groups are one-to-one (e.g. DISO to DISO and CHEM to CHEM), then it is considered homogeneous. We report relation prediction metrics for each of the 6 groups of relation types for RotatE and ComplEx in Table~\ref{tab:relation_categories}. 

We see that RotatE gives impressive relation prediction performance for all groups except for many-to-many-homogeneous, a seemingly challenging group of relations containing ambiguous and synonymous relation types, e.g. \verb possibly_equivalent_to , \verb same_as , \verb refers_to , \verb isa . The full list of M-M-hom relations are shown in the Appendix. In contrast, ComplEx struggles with a wider array of relation types, suggesting that it is generally less able to model different types than RotatE. The last two rows under each model show per-relation results for the causative relationships mentioned previously: \verb cause_of  and \verb causative_agent_of . RotatE again shows significantly better results compared to ComplEx, in line with its theoretically superior representation capacity \cite{sun2018rotate}.

 \section{Discussion}
Based on our findings, we recommend the use of KGE models to leverage the multi-relational nature of knowledge graphs for learning biomedical concept and relation embeddings; and of appropriate evaluation tasks such as link prediction, entity classification and relation prediction for fair comparison across models. We also encourage analysis beyond standard validation metrics, e.g. visualization, examining model predictions, reporting metrics for different relation groupings and devising problem or domain-specific validation tasks. A further promising evaluation task is the triple prediction proposed in \cite{allen_understanding_2019}, which we leave for future work. A more ideal way to assess concept embeddings in biomedical NLP applications and patient-level modeling would be to design a suite of benchmark downstream tasks that incorporate the embeddings, but that warrants a rigorous paper of its own and is left for future work.

We believe this paper serves the biomedical NLP community as an introduction to KGEs and their evaluation and analyses, and also the KGE community by providing a potential standard benchmark dataset with real-world relevance. 

\begin{table}[!tp]
    \centering
    \begin{tabular}{c|c|c|c|r}
    \hline
        \textbf{Relation} & \textbf{MRR} & \textbf{H@1} & \textbf{H@10} & \textbf{Count}  \\
        \hline
        \multicolumn{5}{c}{ComplEx}\\\hline
        1-1-hom & .600  & .319  & .944  & 72 \\
        M-M-hom & .605  & .417  & .877  & 29,028 \\
        M-1     & .683  & .557  & .884  & 2,509 \\
        1-M     & .738  & .640  & .916  & 2,497 \\
        1-1     & .889  & .817  & .995  & 420 \\
        M-M     & .867  & .819  & .941  & 15,044 \\\hline
        Co & .706 & .662 & .779 & 145 \\
        CA & .857 & .822 & .908 & 303 \\
        \hline
        \multicolumn{5}{c}{RotatE}\\\hline
        M-M-hom & .784  & .718    & .934  & 29,028 \\
        M-M     & .973  & .944    & .992  & 15,044 \\
        M-1     & .971  & .945    & .998  & 2,509  \\
        1-M     & .975  & .953    & .998  & 2,497  \\
        1-1     & .985  & .959    & 1.   & 420 \\
        1-1-hom & .972  & .976    & 1.   & 72 \\\hline
        Co & .803 & .738 & .890  & 145 \\
        CA & .996 & .993 & 1.   & 303 \\
        \hline
    \end{tabular}
    \caption{Relation prediction results for RotatE and ComplEx by category of relation type (last two rows relate to causative relation types).}
    \label{tab:relation_categories}
\end{table}

\section{Conclusion and Future Work}

We present results from applying 5 leading KGE models to the SNOMED-CT knowledge graph and compare them to related work through visualizations and evaluation tasks, making a case for the importance of using models that leverage the multi-relation nature of knowledge graphs for learning biomedical knowledge representation. We discuss best practices for working with biomedical knowledge graphs and evaluating the embeddings learned from them, proposing link prediction, entity classification, and relation prediction as standard evaluation tasks. We encourage researchers to engage in further validation through visualizations, error analyses based on model predictions, examining stratified metrics, and devising domain-specific tasks that can assess the usefulness of the embeddings for a given application domain. 

There are several immediate avenues of future work. While we focus on the SNOMED-CT dataset and the KGE models implemented in GraphVite, other biomedical terminologies such as the Gene Ontology \cite{10.1093/nar/gky1055} and RxNorm \cite{rxnorm} could be explored and more recent KGE models, e.g. TuckER \cite{balazevic2019tucker} and MuRP \cite{balazevic2019multi}, applied. 
Additional sources of information could also potentially be incorporated, such as textual descriptions of entities and relations. In preliminary experiments, we initialized entity and relation embeddings with the embeddings of their textual descriptors extracted using Clinical Bert \cite{alsentzer-etal-2019-publicly}, but it did not yield gains. This may suggest that the concept and language spaces are substantially different and strategies to jointly train with linguistic and knowledge graph information require further study. Other sources of information include entity types (e.g. UMLS semantic type) and paths, or multi-hop generalizations of the 1-hop relations (triples) typically used in KGE models \cite{guu-etal-2015-traversing}. Notably, CoKE trains contextual knowledge graph embeddings using path-level information under an adapted version of the BERT training paradigm \cite{wang2019:coke}. 

Lastly, the usefulness of biomedical knowledge graph embeddings should be investigated in downstream applications in biomedical NLP such as information extraction, concept normalization and entity linking, computational fact checking, question answering, summarization, and patient trajectory modeling. In particular, entity linkers act as a bottleneck between text and concept spaces, and leveraging KGEs could help develop sophisticated tools to parse existing biomedical and clinical text datasets for concept-level annotations and additional insights. Well performing entity linkers may then enable training knowledge-grounded large-scale language models like KnowBert \cite{Peters2019KnowledgeEC}. Overall, methods for learning and incorporating domain-specific knowledge representation are still at an early stage and further discussions are needed.


\bibliography{snomed_kge}

\begin{thebibliography}{26}
\expandafter\ifx\csname natexlab\endcsname\relax\def\natexlab#1{#1}\fi

\bibitem[{Agarwal et~al.(2019)Agarwal, Eftimov, Addanki, Choudhury, Tamang, and
  Rallo}]{agarwal_snomed2vec_2019}
Khushbu Agarwal, Tome Eftimov, Raghavendra Addanki, Sutanay Choudhury, Suzanne
  Tamang, and Robert Rallo. 2019.
\newblock \href {http://arxiv.org/abs/1907.08650} {{Snomed2Vec}: {Random}
  {Walk} and {Poincare} {Embeddings} of a {Clinical} {Knowledge} {Base} for
  {Healthcare} {Analytics}}.
\newblock \emph{arXiv:1907.08650 [cs, stat]}.
\newblock ArXiv: 1907.08650.

\bibitem[{Allen et~al.(2019)Allen, Balazevic, and
  Hospedales}]{allen_understanding_2019}
Carl Allen, Ivana Balazevic, and Timothy~M. Hospedales. 2019.
\newblock \href {http://arxiv.org/abs/1909.11611} {On {Understanding}
  {Knowledge} {Graph} {Representation}}.
\newblock \emph{arXiv:1909.11611 [cs, stat]}.

\bibitem[{Alsentzer et~al.(2019)Alsentzer, Murphy, Boag, Weng, Jin, Naumann,
  and McDermott}]{alsentzer-etal-2019-publicly}
Emily Alsentzer, John Murphy, William Boag, Wei-Hung Weng, Di~Jin, Tristan
  Naumann, and Matthew McDermott. 2019.
\newblock \href {https://doi.org/10.18653/v1/W19-1909} {Publicly available
  clinical {BERT} embeddings}.
\newblock In \emph{Proceedings of the 2nd Clinical Natural Language Processing
  Workshop}, pages 72--78, Minneapolis, Minnesota, USA. Association for
  Computational Linguistics.

\bibitem[{Bala\v{z}evi\'c et~al.(2019)Bala\v{z}evi\'c, Allen, and
  Hospedales}]{balazevic2019tucker}
Ivana Bala\v{z}evi\'c, Carl Allen, and Timothy~M Hospedales. 2019.
\newblock Tucker: Tensor factorization for knowledge graph completion.
\newblock In \emph{Empirical Methods in Natural Language Processing}.

\bibitem[{Bala{\v{z}}evi{\'c} et~al.(2019)Bala{\v{z}}evi{\'c}, Allen, and
  Hospedales}]{balazevic2019multi}
Ivana Bala{\v{z}}evi{\'c}, Carl Allen, and Timothy Hospedales. 2019.
\newblock Multi-relational poincar$\backslash$'e graph embeddings.
\newblock In \emph{Advances in Neural Information Processing Systems}.

\bibitem[{Beam et~al.(2019)Beam, Kompa, Schmaltz, Fried, Weber, Palmer, Shi,
  Cai, and Kohane}]{beam_clinical_2019}
Andrew~L. Beam, Benjamin Kompa, Allen Schmaltz, Inbar Fried, Griffin Weber,
  Nathan~P. Palmer, Xu~Shi, Tianxi Cai, and Isaac~S. Kohane. 2019.
\newblock \href {http://arxiv.org/abs/1804.01486} {Clinical {Concept}
  {Embeddings} {Learned} from {Massive} {Sources} of {Multimodal} {Medical}
  {Data}}.
\newblock \emph{arXiv:1804.01486 [cs, stat]}.
\newblock ArXiv: 1804.01486.

\bibitem[{Bodenreider(2004)}]{bodenreider_unified_2004}
Olivier Bodenreider. 2004.
\newblock \href {https://doi.org/10.1093/nar/gkh061} {The {Unified} {Medical}
  {Language} {System} ({UMLS}): integrating biomedical terminology}.
\newblock \emph{Nucleic Acids Research}, 32(90001):267D--270.

\bibitem[{Bordes et~al.(2013)Bordes, Usunier, Garcia-Duran, Weston, and
  Yakhnenko}]{transe}
Antoine Bordes, Nicolas Usunier, Alberto Garcia-Duran, Jason Weston, and Oksana
  Yakhnenko. 2013.
\newblock \href
  {http://papers.nips.cc/paper/5071-translating-embeddings-for-modeling-multi-relational-data.pdf}
  {Translating embeddings for modeling multi-relational data}.
\newblock In C.~J.~C. Burges, L.~Bottou, M.~Welling, Z.~Ghahramani, and K.~Q.
  Weinberger, editors, \emph{Advances in Neural Information Processing Systems
  26}, pages 2787--2795. Curran Associates, Inc.

\bibitem[{Choi et~al.(2016)Choi, Bahadori, Searles, Coffey, Thompson, Bost,
  Tejedor-Sojo, and Sun}]{choi_multi-layer_2016}
Edward Choi, Mohammad~Taha Bahadori, Elizabeth Searles, Catherine Coffey,
  Michael Thompson, James Bost, Javier Tejedor-Sojo, and Jimeng Sun. 2016.
\newblock \href {https://doi.org/10.1145/2939672.2939823} {Multi-layer
  {Representation} {Learning} for {Medical} {Concepts}}.
\newblock In \emph{Proceedings of the 22nd {ACM} {SIGKDD} {International}
  {Conference} on {Knowledge} {Discovery} and {Data} {Mining} - {KDD} '16},
  pages 1495--1504, San Francisco, California, USA. ACM Press.

\bibitem[{Dettmers et~al.(2018)Dettmers, Pasquale, Pontus, and
  Riedel}]{dettmers2018conve}
Tim Dettmers, Minervini Pasquale, Stenetorp Pontus, and Sebastian Riedel. 2018.
\newblock \href {https://arxiv.org/abs/1707.01476} {Convolutional 2d knowledge
  graph embeddings}.
\newblock In \emph{Proceedings of the 32th AAAI Conference on Artificial
  Intelligence}, pages 1811--1818.

\bibitem[{Grover and Leskovec(2016)}]{node2vec-grover}
Aditya Grover and Jure Leskovec. 2016.
\newblock node2vec: Scalable feature learning for networks.
\newblock In \emph{Proceedings of the 22nd ACM SIGKDD International Conference
  on Knowledge Discovery and Data Mining}.

\bibitem[{Guu et~al.(2015)Guu, Miller, and Liang}]{guu-etal-2015-traversing}
Kelvin Guu, John Miller, and Percy Liang. 2015.
\newblock \href {https://doi.org/10.18653/v1/D15-1038} {Traversing knowledge
  graphs in vector space}.
\newblock In \emph{Proceedings of the 2015 Conference on Empirical Methods in
  Natural Language Processing}, pages 318--327, Lisbon, Portugal. Association
  for Computational Linguistics.

\bibitem[{Hitchcock(1927)}]{hitchcock}
Frank~L. Hitchcock. 1927.
\newblock \href {https://doi.org/10.1002/sapm192761164} {The expression of a
  tensor or a polyadic as a sum of products}.
\newblock \emph{Journal of Mathematics and Physics}, 6(1-4):164--189.

\bibitem[{Kazemi and Poole(2018)}]{kazemi}
Seyed Kazemi and David Poole. 2018.
\newblock Simple embedding for link prediction in knowledge graphs.
\newblock In \emph{Advances in Neural Information Processing Systems 32}.

\bibitem[{Kotitsas et~al.(2019)Kotitsas, Pappas, Androutsopoulos, McDonald, and
  Apidianaki}]{kotitsas_embedding_2019}
Sotiris Kotitsas, Dimitris Pappas, Ion Androutsopoulos, Ryan McDonald, and
  Marianna Apidianaki. 2019.
\newblock \href {https://doi.org/10.18653/v1/W19-5032} {Embedding {Biomedical}
  {Ontologies} by {Jointly} {Encoding} {Network} {Structure} and {Textual}
  {Node} {Descriptors}}.
\newblock In \emph{Proceedings of the 18th {BioNLP} {Workshop} and {Shared}
  {Task}}, pages 298--308, Florence, Italy. Association for Computational
  Linguistics.

\bibitem[{Mikolov et~al.(2013)Mikolov, Sutskever, Chen, Corrado, and
  Dean}]{mikolov_2013}
Tomas Mikolov, Ilya Sutskever, Kai Chen, Greg~S Corrado, and Jeff Dean. 2013.
\newblock \href
  {http://papers.nips.cc/paper/5021-distributed-representations-of-words-and-phrases-and-their-compositionality.pdf}
  {Distributed representations of words and phrases and their
  compositionality}.
\newblock In C.~J.~C. Burges, L.~Bottou, M.~Welling, Z.~Ghahramani, and K.~Q.
  Weinberger, editors, \emph{Advances in Neural Information Processing Systems
  26}, pages 3111--3119. Curran Associates, Inc.

\bibitem[{Nelson et~al.(2011)Nelson, Zeng, Kilbourne, Powell, and
  Moore}]{rxnorm}
Stuart Nelson, Kelly Zeng, John Kilbourne, Tammy Powell, and Robin Moore. 2011.
\newblock \href {https://doi.org/10.1136/amiajnl-2011-000116} {{Normalized
  names for clinical drugs: RxNorm at 6 years}}.
\newblock \emph{JAMIA}, 18(4):441--448.

\bibitem[{Nickel et~al.(2011)Nickel, Tresp, and Kriegal}]{rescal2011}
Maximilian Nickel, Volker Tresp, and Hans-Peter Kriegal. 2011.
\newblock A three-way model for collective learning on multi-relational data.
\newblock In \emph{Proceedings of the 28th International Conference on Machine
  Learning}, pages 809--816.

\bibitem[{Peters et~al.(2019)Peters, Neumann, Logan, Schwartz, Joshi, Singh,
  and Smith}]{Peters2019KnowledgeEC}
Matthew~E. Peters, Mark Neumann, Robert~L Logan, Roy Schwartz, Vidur Joshi,
  Sameer Singh, and Noah~A. Smith. 2019.
\newblock Knowledge enhanced contextual word representations.
\newblock In \emph{EMNLP}.

\bibitem[{Sun et~al.(2019)Sun, Deng, Nie, and Tang}]{sun2018rotate}
Zhiqing Sun, Zhi-Hong Deng, Jian-Yun Nie, and Jian Tang. 2019.
\newblock \href {https://openreview.net/forum?id=HkgEQnRqYQ} {Rotate: Knowledge
  graph embedding by relational rotation in complex space}.
\newblock In \emph{International Conference on Learning Representations}.

\bibitem[{Tang et~al.(2016)Tang, Liu, Zhang, and Mei}]{tang2016visualizing}
Jian Tang, Jingzhou Liu, Ming Zhang, and Qiaozhu Mei. 2016.
\newblock Visualizing large-scale and high-dimensional data.
\newblock In \emph{Proceedings of the 25th International Conference on World
  Wide Web}, pages 287--297. International World Wide Web Conferences Steering
  Committee.

\bibitem[{The Gene Ontology Consortium(2018)}]{10.1093/nar/gky1055}
The Gene Ontology Consortium. 2018.
\newblock \href {https://doi.org/10.1093/nar/gky1055} {{The Gene Ontology
  Resource: 20 years and still GOing strong}}.
\newblock \emph{Nucleic Acids Research}, 47(D1):D330--D338.

\bibitem[{Trouillon et~al.(2016)Trouillon, Welbl, Riedel, Gaussier, and
  Bouchard}]{trouillon2016complex}
Th\'eo Trouillon, Johannes Welbl, Sebastian Riedel, \'Eric Gaussier, and
  Guillaume Bouchard. 2016.
\newblock {Complex embeddings for simple link prediction}.
\newblock In \emph{International Conference on Machine Learning (ICML)},
  volume~48, pages 2071--2080.

\bibitem[{Wang et~al.(2019)Wang, Huang, Wang, Dai, Jiang, Liu, Lyu, and
  Wu}]{wang2019:coke}
Quan Wang, Pingping Huang, Haifeng Wang, Songtai Dai, Wenbin Jiang, Jing Liu,
  Yajuan Lyu, and Hua Wu. 2019.
\newblock Coke: Contextualized knowledge graph embedding.
\newblock \emph{arXiv:1911.02168}.

\bibitem[{Yang et~al.(2015)Yang, Yih, He, Gao, and Deng}]{yang_embedding_2015}
Bishan Yang, Scott Wen-tau Yih, Xiaodong He, Jianfeng Gao, and Li~Deng. 2015.
\newblock \href
  {https://www.microsoft.com/en-us/research/publication/embedding-entities-and-relations-for-learning-and-inference-in-knowledge-bases/}
  {Embedding {Entities} and {Relations} for {Learning} and {Inference} in
  {Knowledge} {Bases}}.
\newblock In \emph{Proceedings of the {International} {Conference} on
  {Learning} {Representations} ({ICLR}) 2015}.

\bibitem[{Zhu et~al.(2019)Zhu, Xu, Qu, and Tang}]{zhu2019graphvite}
Zhaocheng Zhu, Shizhen Xu, Meng Qu, and Jian Tang. 2019.
\newblock Graphvite: A high-performance cpu-gpu hybrid system for node
  embedding.
\newblock In \emph{The World Wide Web Conference}, pages 2494--2504. ACM.

\end{thebibliography}
\bibliographystyle{acl_natbib}

\end{document}